\documentclass[final]{cvpr}

\usepackage{times}
\usepackage{epsfig}
\usepackage{graphicx}
\usepackage{amsmath}
\usepackage{amssymb}
\usepackage{xcolor}
\usepackage{arydshln}
\def\mybox#1{\leavevmode \setbox0=\hbox{\framebox{#1}}%
   \dimen0=\wd0 \edef\posxA{\expandafter\ignorept\the\dimen0 \space}%
   \hbox{\kern3pt\pdfliteral{q .8 .8 1 rg .8 .8 1 RG .9963 0 0 .9963 0 0 cm 1 j 1 J 6 w
                             0 0 m 0 5 l \posxA 5 l \posxA 0 l 0 0 l B Q}%
         \box0 \kern3pt}%
}
{\lccode`\?=`\p \lccode`\!=`\t  \lowercase{\gdef\ignorept#1?!{#1}}}

\usepackage{hyperref}
\hypersetup{pagebackref=true,breaklinks=true,colorlinks,bookmarks=false}

\def\cvprPaperID{2} 
\def\confYear{CVPR 2021}

\begin{document}

\title{Finding Facial Forgery Artifacts with Parts-Based Detectors}

\author{Steven Schwarcz\\
University of Maryland\\
College Park, MD\\
{\tt\small schwarcz@umiacs.umd.edu}
\and
Rama Chellappa\\
Johns Hopkins University\\
Baltimore, MD\\
{\tt\small rchella4@jhu.edu}
}

\maketitle

\begin{abstract}
Manipulated videos, especially those where the identity of an individual has been modified using deep neural networks, are becoming an increasingly relevant threat in the modern day. In this paper, we seek to develop a generalizable, explainable solution to detecting these manipulated videos. To achieve this, we design a series of forgery detection systems that each focus on one individual part of the face. These parts-based detection systems, which can be combined and used together in a single architecture, meet all of our desired criteria - they generalize effectively between datasets and give us valuable insights into what the network is looking at when making its decision. We thus use these detectors to perform detailed empirical analysis on the FaceForensics++, Celeb-DF, and Facebook Deepfake Detection Challenge datasets, examining not just what the detectors find but also collecting and analyzing useful related statistics on the datasets themselves.
\end{abstract}

\section{Introduction}

The past few years have seen a sharp rise in the availability of technology for creating and distributing digital forgeries, specifically those where one individual's identity is replaced with that of a another. These forgeries, known as ``Deepfakes,'' represent an important and growing threat to information integrity on the web. A variety of techniques have been proposed to combat the risks inherent in the widespread availability of this technology, many of which are known to perform very well on the datasets on which they are trained, but perform worse when they are applied to different datasets, particularly those which generate fakes using different algorithms.

\begin{figure}[]
\begin{center}
   \includegraphics[width=0.95\linewidth]{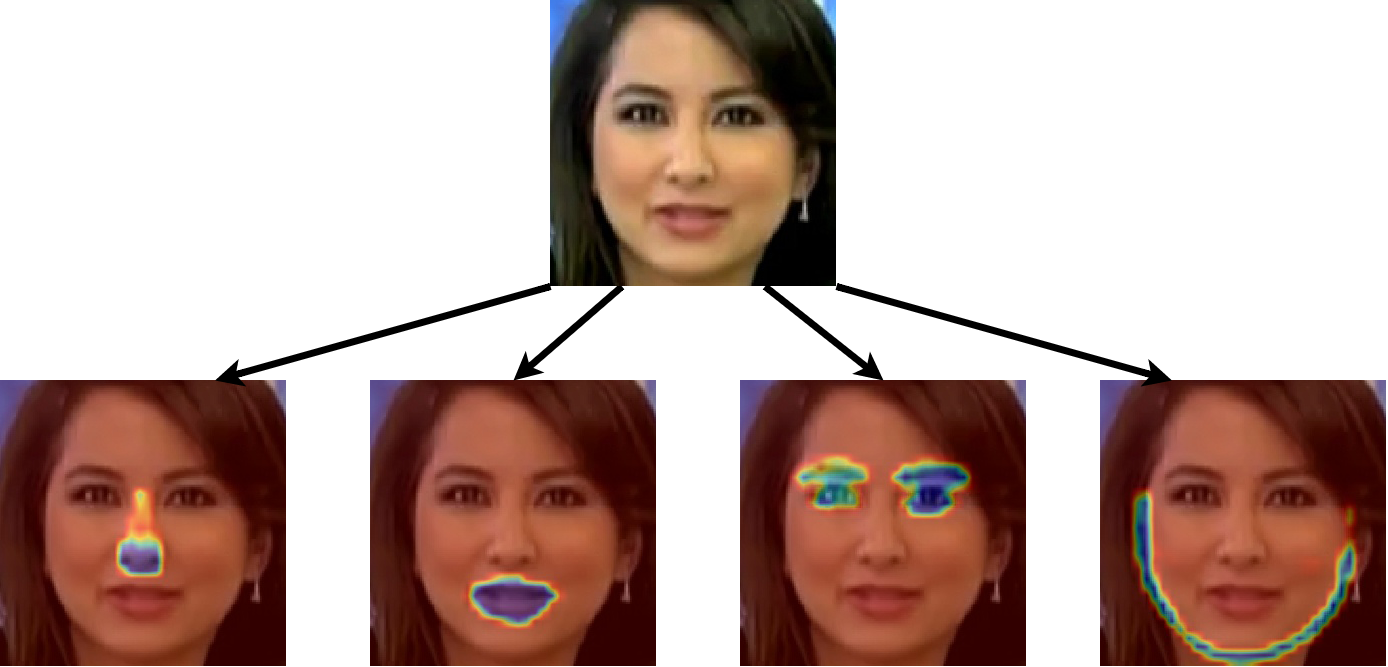}
\end{center}
   \caption{In order to detect manipulated images in an explainable way, we divide the face into four regions - nose, mouth, eyes and chin/jawline - and train separate classifiers for each region. The resultant classifiers generalize well and provide valuable insights into the image manipulation process.}
\label{fig:intro}
\end{figure}

For this reason, it is vital that video forgery detection systems be able to generalize effectively to novel datasets. Improving performance in this regard requires a strong understanding of both the detectors used and the data itself. To this end, a lot of approaches have been explored \cite{Li2020FaceXF,Chai2020,Li2019ExposingDV}.

In this paper, we seek to add one more approach to this list, analyzing the problem of digital forgery detection from another angle. Rather than focus on a single architecture for forgery detection, we break the problem down into smaller pieces, dividing the face into several distinct regions of interest and training separate classifiers for each one. Specifically, we divide the face into four main regions - the nose, mouth, eyes and chin - and learn separate detectors for the individual artifacts left behind in each region. 

Despite the fact that these classifiers are designed to restrict their attention only to their specific assigned regions, some of them still generalize quite well. This suggests that the regions we are training on may contain powerful clues for forgery detection, powerful enough to let these restricted classifiers out-perform more conventional systems.

These parts-based detectors are also fertile ground for deeper empirical analysis. They are explainable because they make decisions based on limited, easily understood criteria, and we can use their strengths or weaknesses to make inferences about the underlying data they are trained and evaluated on. They thus allow us to gain further insights into the comparative structure of different forgery algorithms. 

Using FaceForensics++ \cite{Rossler2019} as our main dataset, our empirical analysis is broken down as follows: first, we train and evaluate the generalizability of parts-based detectors for each of the four regions of the face that we designate - nose, mouth, eyes and chin/jawline.  We then perform the same experiments with a combined parts-detector, which uses multiple branches within the same network to compute parts-based detection separately before recombining. Having trained these parts detectors, we perform comparative analysis of their strengths and weaknesses, looking at both the performance of our detectors and the statistics of altered regions in manipulated images in an effort to develop new insights into the problem of forgery detection. We finish with cross-dataset analysis, comparing generalization performance not just between different algorithms of the FaceForensics++ dataset but also across datasets, performing transfer experiments on the Celeb-DF \cite{celebdf} and Facebook Deepfake Detection Challenge \cite{dfdc} datasets.

\section{Related Works}

\subsection{Deepfake Algorithms}

Computers have long been used to generate forged imagery, but until recently most quality digital forgeries would need to be carefully designed by hand. However, since the invention and widespread use of Generative Adversarial Networks (GANs) \cite{gan}, it has become considerably easier to use computers to convincingly modify images. In this paper, we are particularly concerned with digital forgery techniques that use deep learning or other modern techniques to transfer the face of one individual to another.

These forgeries, colloquially known as ``Deepfakes,'' can be generated in a variety of ways. Some of these methods predate the invention of GANs, such as \cite{faceswap}, which performs swaps by identifying landmarks on faces, and then warping one individual's face onto the other. Still, many Deepfake methods have GANs at their core. FSGAN \cite{fsgan} uses a GAN to reenact one person's pose with another's face, while the NTH model \cite{9009591} produces Deepfakes by pre-training meta-variables and swapping faces in a few-shot setting. Similarly, techniques like StyleGAN \cite{stylegan} allow Deepfake generation by projecting one image into the latent space of another.

These techniques are especially easy to use because many of them have been made available as software packages. Software such as DeepFaceLab \cite{perov2020deepfacelab} and FaceSwap \footnote{https://github.com/deepfakes/faceswap} are easy to download and use, making Deepfakes available to many people outside the research community.

\subsection{Forgery/Deepfake Detection}

As techniques for generating forged images have improved, so too have techniques for identifying them. Some techniques identified forgery by looking at metadata \cite{Huh_2018_ECCV} or other low level artifacts. For instance, techniques such as \cite{1381775,10.1145/1288869.1288876,1511009,Li2019ExposingDV,Li2020FaceXF} try to identify forgeries by looking for specific artifacts, e.g.\ by searching for repeated regions, or stitching or warping artifacts in transferring one face to another body. Patch-based techniques are also popular \cite{8014963,8267647}; for instance, \cite{Chai2020} uses patch-based techniques to analyze and improve the generalizability of video forgery detection systems.

Many techniques take advantage of the strength of deep networks directly, by training image classifiers with various architectures to identify forgeries, either through RGB \cite{Bayar,mesonet,Mo2018,cozzolino2017recasting} or optical flow \cite{9022558}. Our work continues in the same vein as many of these other works, taking key intuitions from other methods. In particular we take inspiration from the patch-based approach of \cite{Chai2020}, extending their patch-based approach to train specific parts-based detectors.

There have also been many new datasets introduced to evaluate forgery detection systems. These range from the smaller but varied Deepfake TIMIT \cite{Korshunov} and FaceForensics++ \cite{Rossler2019} datasets, to the larger, more recent Facebook Deepfake Detection Challenge (DFDC) \cite{dfdc} and Celeb-DF  \cite{celebdf} datasets, the last three of which we use in this work.

\section{Method}

We seek to develop an explainable infrastructure for detecting manipulated images and videos. Below, we explain our approach for building separate part-based classifiers for each region of the face. These classifiers - which must make their decisions by looking at only a limited portion of the image - allow us to get the explainable detectors we desire, which will be critical in performing our analysis.

\subsection{Creating the Parts Masks \label{sec:mask_gen}}

\begin{figure}[]
\begin{center}
   \includegraphics[width=0.85\linewidth]{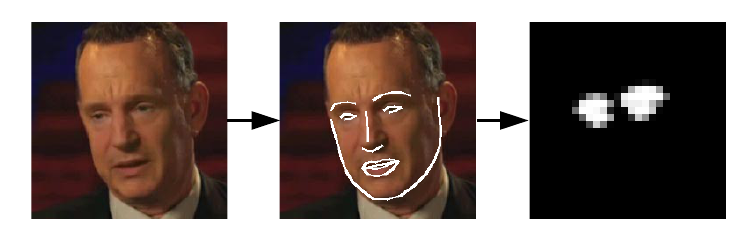}
\end{center}
   \caption{Mask generation pipeline. First parts are detected using dlib, then the masks are created from the convex hulls of the detected regions. Finally, masks are post-processed with morphological  dilations and gaussian blur.}
\label{fig:mask_gen}
\end{figure}

In order to train parts-specific detectors, we must create individual masks for each region of the face so that they can be used during training. Figure \ref{fig:mask_gen} illustrates this straight-forward process. We first acquire facial landmarks using the dlib \cite{dlib09} software library. These detected landmarks are subsequently grouped into four categories: nose, mouth, eyes, and chin/jawline. To convert these landmarks into masks, we simply take the convex hulls of each region separately, except for the chin which we keep as a series of line segments. We then perform 8 iterations of morphological dilation and apply a gaussian filter to create the final high resolution masks. Since these masks will be used to train low-resolution filters within the network described below, the final step in our mask generation pipeline is to downsample the mask to a resolution that matches the maps produced by our network.

\subsection{Parts Detection \label{sec:arch_single}}

\textbf{Single Part-Based Classifier} We first describe our detector in the context of a single fixed facial region $R$, such as the mouth or eyes, which we will refer to as an $R$-based detector. Since we are only interested in a small portion of the face at a time, we opt to use a patch-based detector with a small receptive field. Thus, as a backbone for our network we use a standard Xception \cite{xception} neural network, truncated after the second Xception block and given a single channel of output using a $1 \times 1$ convolution layer. This is the same architecture used in \cite{Chai2020}, and we adopt it in part because the authors of \cite{Chai2020} demonstrated that a truncated Xception network generalizes more effectively than the full, deeper network.

Unlike \cite{Chai2020}, however, we do not directly classify from the output of the truncated Xception network. Instead, in the case of fake images, we use these truncated outputs in order to learn a mask for the region $R$. Specifically, if $x$ is our input image, $f(x)$ is the output of our truncated neural network, and $M_R$ is the binary mask for region $R$ constructed as described in Section \ref{sec:mask_gen}, then we train an $R$-based classifier by minimizing standard binary cross-entropy loss:

\begin{equation} \label{eq:mask_loss}
\begin{split}
\mathcal{L}_{M}(x, M_R) = & - \sum_{i, j} M_{R,ij}\log(\sigma(f(x)_{ij})) \\
& \quad -  \sum_{i, j} (1 - M_{R,ij}) \log(1 - \sigma(f(x)_{ij}))
\end{split}
\end{equation}

\noindent where $i$ and $j$ are the horizontal and vertical indices of a given map, and $\sigma(\cdot)$ is the sigmoid function. 

If the image being classified is real, then $M_R$ in equation \ref{eq:mask_loss} is set to all zeros, teaching the network not to activate when given a real image. Thus, $\mathcal{L}_M$ teaches the network to only look for artifacts of forgery in the specific region targeted by $M_R$. 

Once $f(x)$ has been generated, we produce a final classification label $\hat{y}$ by performing average pooling over $f(x)$ and feeding the result through a classification layer. We denote this operation as $\hat{y} = g(f(x))$. $\hat{y}$ is trained by minimizing binary cross entropy as well, this time over a single value only:

\begin{equation} \label{eq:class_loss}
\begin{split}
\mathcal{L}_{C}(x) = - y\log(\sigma(\hat{y}))  -  (1 - y) \log(1 - \sigma(\hat{y}))
\end{split}
\end{equation}

\noindent where $y$ is the ground truth label of the image. The fact that the network must make its prediction solely from the pooled results of the part detector ensures that only the activations of the part detector are used in the final classification.

Thus, for an $R$-based classifier over a single region, our final loss is:

\begin{equation} 
\mathcal{L}_{R}(x) = \mathcal{L}_{C}(x) + \lambda \mathcal{L}_M(x, M_R)
\end{equation}

\noindent where $\lambda$ is a learning weight parameter set to $10$ in all of our experiments.

We take a moment here to remark on the small receptive field of our classifier. \cite{Chai2020} argued that this smaller receptive field helps with generalization, but it also provides another advantage in our analysis. Namely, it ensures that our network is only looking at the regions we want it to. If the receptive fields were large enough to see the whole image, then it would conceivably be possible that the network might perform classification in an implicit two step process that bypasses looking for artifacts in its  assigned region: first identify if the image is real or manipulated, then detect the appropriate part. With a small receptive field, though, the network can make its decision only by looking in the  area local to whichever region it is focused on classifying. Thus, in all our following analysis, we can be confident that the classifier is making its decision solely from observing the specific region we have trained it to examine.

\textbf{Multiple Parts-Based Classifier} In addition to performing experiments with single-parts based classifiers, we also perform analysis on detectors that combine multiple regions. Since we do not wish to train multiple networks from scratch in order to perform this recombination step, we seek a way to reuse as much computation as possible without having the networks interfere with each other. 

\begin{figure}[h]
\begin{center}
   \includegraphics[width=0.95\linewidth]{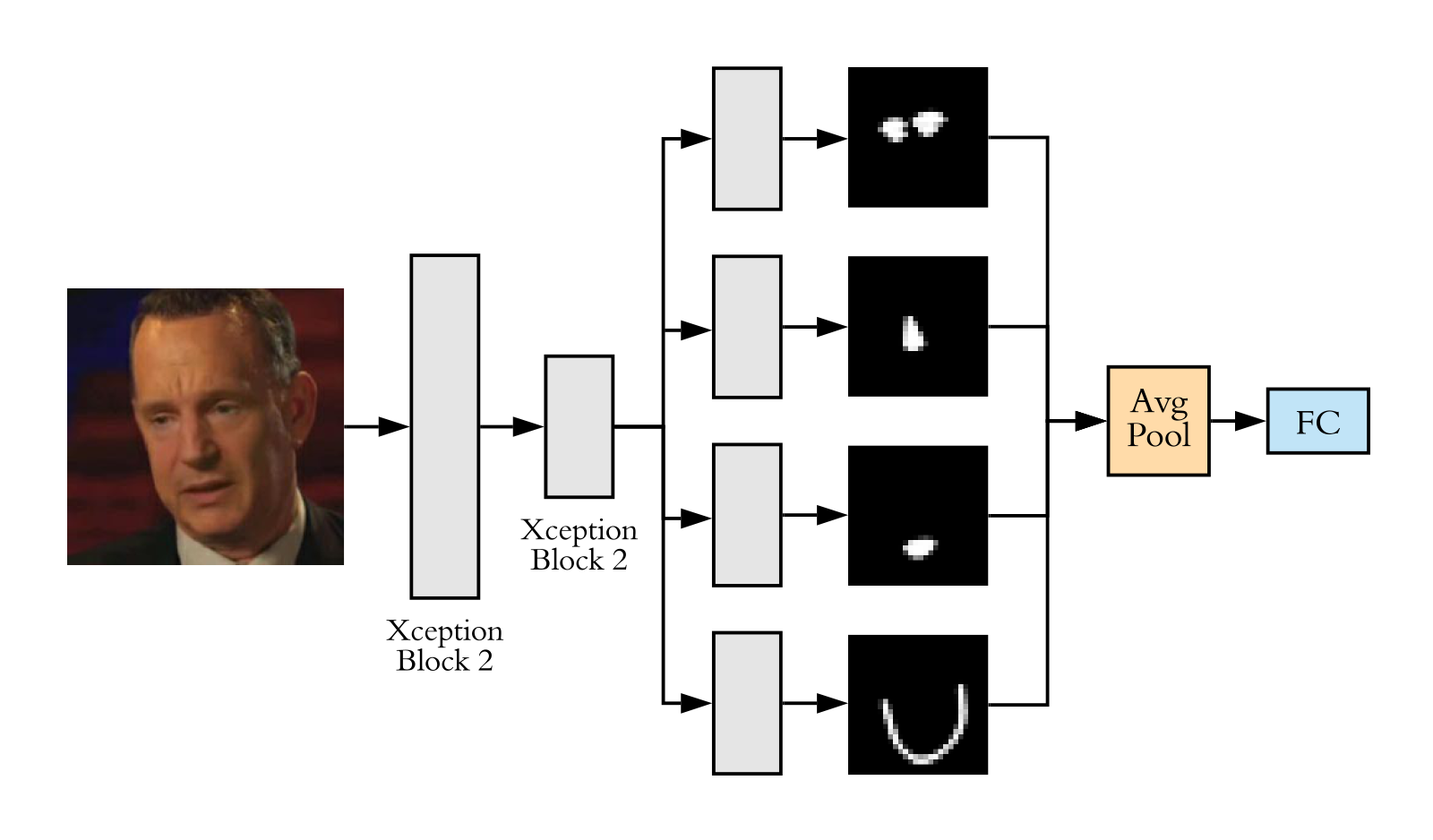}
\end{center}
   \caption{Overview of the architecture used in our experiments. We truncate the Xception architecture after two blocks, then divide the network into separate branches for each part of the face, learning a separate parts mask for each branch. We then concatenate and pool all of the masks into a single value. Finally, we feed this into a fully connected layer to make a prediction.}
\label{fig:arch}
\end{figure}

Our proposed solution to this problem is illustrated in Figure \ref{fig:arch}. When using multiple parts, instead of simply truncating at the second block of the Xception architecture, we add an additional Xception block, identical to the previous one, for each of the four parts of the face we are detecting. None of these additional blocks share weights, thus ensuring that they are each performing their detections separately. We then average the results of all four maps to make our final predictions.

Our final loss for the multiple parts-based classifier is thus:

\begin{equation} 
\mathcal{L}_{multi}(x) = \mathcal{L}_{C}(x) + \lambda \sum_R \mathcal{L}_M(x, M_R)
\end{equation}

\section{Experiments}

\subsection{Data}

The main dataset we use in this paper is FaceForensics++ \cite{Rossler2019}. This dataset consists of 5000 videos broken into three splits: a training split of 3600 videos (720 real and 2880 fake), a validation split of 700 videos (140 real and 560 fake), and a test split of 700 videos (140 real and 560 fake). FaceForensics++ labels the algorithms used to generate each manipulated video, and for each real video fakes are designed using one of the the following algorithms: FaceSwap\footnote{https://github.com/MarekKowalski/FaceSwap} (FS), Deepfakes\footnote{https://github.com/deepfakes/faceswap} (DF), Face2Face \cite{face2face} (F2F), and NeuralTextures \cite{neuraltextures} (NT). For the bulk of our experiments, we will train a system on one of these algorithmic splits, and evaluate on the other three.

For experiments measuring the transfer between datasets, there are two other datasets we use. The first of these is the Celeb-DF v2 dataset \cite{celebdf}, which contains 6229 videos, of which 518 are in the test split that we use in our experiments.  These videos were generated using an improved version of the standard Deepfake synthesis algorithm from 590 different YouTube videos of celebrities.

The second is the Facebook Deepfake Detection Challenge dataset \cite{dfdc}, which at the time of this writing is the largest publicly-available Deepfake dataset collected.  We will only be performing our evaluation on the publicly available test set of 5000 videos, which were created using the algorithms DFAE \cite{perov2020deepfacelab}, MM/NN Face Swap \cite{faceswap}, NTH \cite{9009591}, and FSGAN \cite{fsgan}. 

\subsection{Implementation Details}

For all of our experiments, we use an Xception backbone \cite{xception}, pretrained on ImageNet \cite{imagenet_cvpr09}. The networks are trained on 2 Nvidia GeForce RTX 2080 Ti GPUs for 40,000 steps each with Adam \cite{adam} using a batch size of 128, a $\beta_1$ value of $0.928$, weight decay of $0.00005$ and an initial learning rate of $0.0001$. Every 10000 steps of training, the learning rate is reduced by a factor of 10. All of our code is written in the Tensorflow \cite{tensorflow2015} python library for Deep Learning. 

For every video in each dataset, we randomly select 40 frames to use for training. For all images, we use the Dual Shot Face Detector (DSFD) \cite{Li2018} method to detect the faces within each frame. We then take a crop of this face and resize it to $288 \times 288$ using nearest neighbor interpolation for all of our Xception training. When training the ResNet50 baselines, we use images of size $224 \times 224$ instead.  No augmentation is used during training or testing, and all images are compressed as high-quality JPEGs before being fed through the system.

\subsection{FaceForensics Generalization}


\begin{table*}[t]
\centering
\begin{tabular}{ll|llllll|llll}
Model            &     & DF                          & F2F                         & FS            & NT            &  &    & DF            & F2F           & FS                          & NT                          \\ \hline
ResNet50         & DF  & {\color[HTML]{9B9B9B} 0.98} & 0.51                        & 0.48          & 0.57          &  & FS & 0.5           & 0.54          & {\color[HTML]{9B9B9B} 0.99} & \textbf{0.49}               \\
ResNet50 Block 2 & DF  & {\color[HTML]{9B9B9B} 0.99} & 0.55                        & 0.47          & 0.68          &  & FS & \textbf{0.54} & 0.65          & {\color[HTML]{9B9B9B} 1}    & 0.42                        \\
Xception         & DF  & {\color[HTML]{9B9B9B} 0.98} & 0.52                        & \textbf{0.49} & 0.61          &  & FS & 0.51          & 0.58          & {\color[HTML]{9B9B9B} 0.99} & 0.5                         \\
Xception Block 2 & DF  & {\color[HTML]{9B9B9B} 0.91} & 0.6                         & 0.43          & 0.79          &  & FS & 0.48          & 0.62          & {\color[HTML]{9B9B9B} 0.92} & 0.29                        \\ \cdashline{1-6} \cdashline{8-12}
Nose             & DF  & {\color[HTML]{9B9B9B} 0.97} & \textbf{0.63}               & 0.33          & 0.81          &  & FS & 0.52          & 0.54          & {\color[HTML]{9B9B9B} 0.99} & 0.41                        \\
Mouth            & DF  & {\color[HTML]{9B9B9B} 0.94} & 0.56                        & 0.48          & 0.76          &  & FS & 0.45          & 0.63          & {\color[HTML]{9B9B9B} 0.96} & 0.22                        \\
Eyes             & DF  & {\color[HTML]{9B9B9B} 0.96} & 0.6                         & 0.4           & 0.84          &  & FS & 0.5           & 0.62          & {\color[HTML]{9B9B9B} 0.97} & 0.38                        \\
Chin             & DF  & {\color[HTML]{9B9B9B} 0.96} & 0.59                        & 0.33          & \textbf{0.85} &  & FS & 0.44          & \textbf{0.73} & {\color[HTML]{9B9B9B} 0.99} & 0.33                        \\
Eyes+Chin        & DF  & {\color[HTML]{9B9B9B} 0.97} & 0.58                        & 0.42          & 0.76          &  & FS & 0.48          & 0.71          & {\color[HTML]{9B9B9B} 0.99} & 0.41                        \\
Combined         & DF  & {\color[HTML]{9B9B9B} 0.97} & 0.62                        & 0.39          & 0.83          &  & FS & 0.52          & 0.56          & {\color[HTML]{9B9B9B} 0.98} & 0.32                        \\ \cline{1-6} \cline{8-12} 
ResNet50         & F2F & 0.57                        & {\color[HTML]{9B9B9B} 0.98} & 0.5           & 0.56          &  & NT & 0.56          & 0.51          & 0.48                        & {\color[HTML]{9B9B9B} 0.94} \\
ResNet50 Block 2 & F2F & 0.66                        & {\color[HTML]{9B9B9B} 0.99} & 0.54          & 0.65          &  & NT & 0.67          & 0.52          & 0.43                        & {\color[HTML]{9B9B9B} 0.98} \\
Xception         & F2F & 0.58                        & {\color[HTML]{9B9B9B} 0.98} & 0.52          & 0.54          &  & NT & 0.59          & 0.6           & 0.5                         & {\color[HTML]{9B9B9B} 1}    \\
Xception Block 2 & F2F & 0.7                         & {\color[HTML]{9B9B9B} 0.94} & 0.64          & 0.74          &  & NT & 0.69          & 0.55          & 0.42                        & {\color[HTML]{9B9B9B} 0.98} \\ \cdashline{1-6} \cdashline{8-12}
Nose             & F2F & 0.65                        & {\color[HTML]{9B9B9B} 0.99} & 0.52          & 0.63          &  & NT & 0.67          & 0.63          & \textbf{0.55}               & {\color[HTML]{9B9B9B} 0.98} \\
Mouth            & F2F & 0.53                        & {\color[HTML]{9B9B9B} 0.98} & 0.65          & 0.52          &  & NT & 0.64          & 0.63          & 0.54                        & {\color[HTML]{9B9B9B} 0.99} \\
Eyes             & F2F & \textbf{0.76}               & {\color[HTML]{9B9B9B} 0.98} & \textbf{0.66} & 0.73          &  & NT & 0.64          & 0.51          & 0.47                        & {\color[HTML]{9B9B9B} 0.99} \\
Chin             & F2F & 0.75                        & {\color[HTML]{9B9B9B} 0.95} & 0.65          & 0.74          &  & NT & \textbf{0.84} & \textbf{0.68} & 0.38                        & {\color[HTML]{9B9B9B} 0.99} \\
Eyes+Chin        & F2F & 0.56                        & {\color[HTML]{9B9B9B} 0.98} & 0.62          & 0.48          &  & NT & 0.77          & 0.61          & 0.47                        & {\color[HTML]{9B9B9B} 0.98} \\
Combined         & F2F & \textbf{0.76}               & {\color[HTML]{9B9B9B} 0.95} & 0.53          & \textbf{0.77} &  & NT & 0.66          & 0.62          & 0.53                        & {\color[HTML]{9B9B9B} 0.98}
\end{tabular}
\caption{AUC for the ROC curves of the parts-based detectors for each of the four parts of the face, as well as for the combined detector. The second column indicates which split of FaceForensics++ was used to train the model, while the other columns show the performance on each of those splits. Baselines are above the dotted lines and our parts-based detectors are below. Best results for each run are in bold, while models evaluated on the same split they were trained with are in grey. The dataset abbreviations are as follows: FaceSwap (FS), Deepfakes (DF), Face2Face (F2F), and NeuralTextures (NT).}
\label{tbl:parts}
\end{table*}

We first explore the effectiveness of each of our parts detectors individually. With this analysis, the purpose is to see what can be achieved with networks trained only to identify fake imagery with respect to a specific location in the image, as described in Section \ref{sec:arch_single}.

Table \ref{tbl:parts} shows our performance on these methods. Here, our parts detectors are trained individually on each of the four algorithm-specific splits of the FaceForensics++ dataset, then evaluated on the other three algorithms. For each system we report the Area Under Curve (AUC) of the Receiver-Operating Characteristic (ROC) curve. We use AUC in part because we found that accuracy can be an unstable metric when measuring forgery detection generalization, since during transfer it is not uncommon for a system to be much better at classifying real images than fake images.

We compare our parts detector to 4 other baseline systems. Two of these systems are the standard Xception \cite{xception} and ResNet50 \cite{resnet} architectures, each pretrained on ImageNet \cite{deng2009imagenet}. For each of these architectures, we also compare to a truncated variant, where we end the architecture after the second block (which is either a ResNet or Xception block) similar to \cite{Chai2020}, and compute our prediction as the average of all the logits. This is effectively the same as training the parts-based classifier without $\mathcal{L}_C$, and assigning $M_R$ to be either all zero or all one if the image is real or fake, respectively.

We find from Table \ref{tbl:parts} that certain parts detectors are superior to others. For instance, the mouth-based detector rarely out-performs the Xception based detectors, generally performing much worse. This can be seen, for instance, when observing the performance in transferring from Face2Face to Deepfakes or NeuralTextures. As we will see in Section \ref{sec:hist}, this may be correlated to different patterns of artifacts in those regions.

The eyes and chin-based detectors, on the other hand, perform considerably better, often out-performing or at least matching the baselines. In some cases, such as with transfer from FaceSwap to Face2Face, or from NeuralTextures to Deepfakes, we see significant improvements in performance from using the chin detector alone. This strong performance suggests that important artifacts of the forgery process are left behind in these regions. 

We observe as well that the nose-based detector falls somewhere in between the others. It often performs poorly, but on certain transfers it performs well, such as when transferring from Deepfakes to Face2Face, or from NeuralTextures to FaceSwap.

The rows marked ``Combined'' show the performance of our four-branch multi-headed parts detector using all branches. In many cases, we see that the combined detector falls somewhere in between the best and worst parts detectors for each given split. 

Since the chin and eyes-based detectors tend to perform better than the other parts, we also include an ``Eyes+Chin'' detector, which uses two branches instead of the four used by the ``Combined'' method. Like the ``Combined'' method, however, we see that combining the detectors does not necessarily greatly improve performance.

\subsubsection{Dataset Parts Breakdown \label{sec:hist}}

\begin{figure}[t]
\begin{center}
   \includegraphics[width=0.9\linewidth]{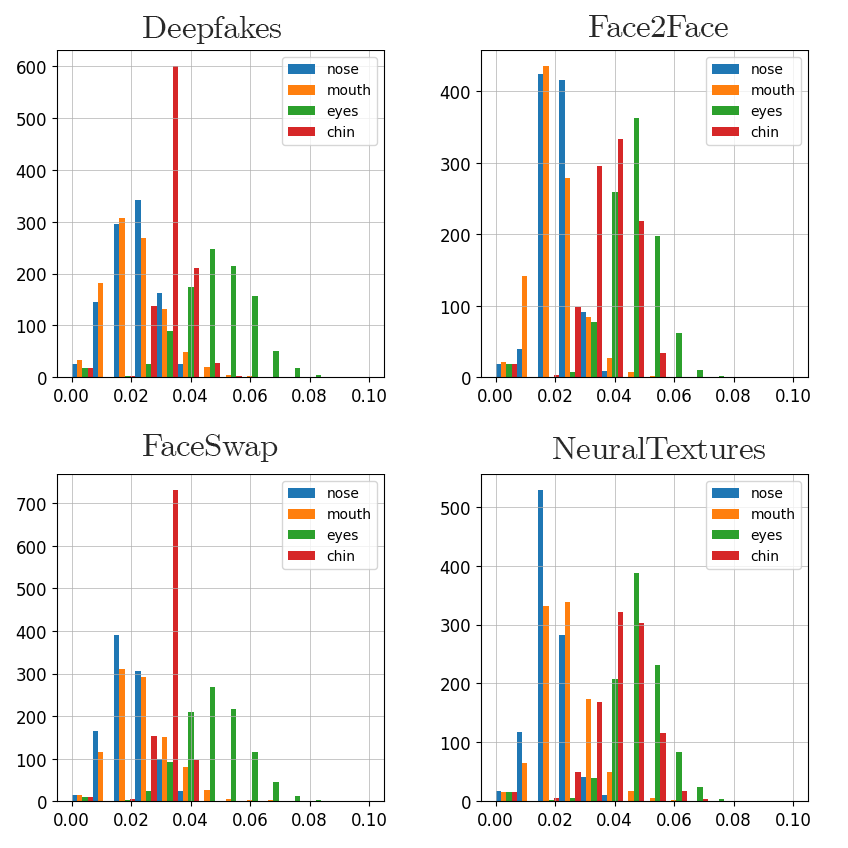}
\end{center}
   \caption{Histograms of absolute pixel difference in each of the four regions of the face used in our analysis, broken down by algorithmic split of the FaceForensics++ dataset.}
\label{fig:hist}
\end{figure}

In order to better make sense of the performance discussed in Table \ref{tbl:parts}, and in an effort to provide some intuition for certain results, we now perform pixel-level analysis on the FaceForensics++ dataset itself. To do this, we first compute, for one frame per video per algorithm split, the absolute difference between the manipulated image and its real counterpart. We then take the same masks $M_R$ used to generate the ground truth for the $R$-based part detectors, and multiply them by this computed absolute difference. Specifically, for each part $R$, given real image $x_r$ and manipulated image $x_f$, we compute a map $D_R$ as 

\begin{equation}
D_{R,ij} = M_{R, ij}\left| x_{r, ij} - x_{f,ij} \right|.
\end{equation}

\noindent In Figure \ref{fig:hist} and Table \ref{tbl:diff}, we provide statistical summaries of the maps $D_R$ for each part $R$ and each data split.

We posit that these statistics tell us something about where to find artifacts left over from the manipulation process. Although these summary statistics leave out a lot of important low-level information, some patterns do emerge when comparing certain results from Table \ref{tbl:parts} to the histograms in Figure \ref{fig:hist}. For instance, the Nose detector performs well on the transfer from NeuralTextures to FaceSwap (.55 vs a .42 truncated Xception baseline and a .5 Xception baseline), and we all see very similar histogram shapes for pixel differences in the nose region on those two datasets.

\begin{table}[t]
\centering
\begin{tabular}{l|llll}
               & Nose   & Mouth  & Eyes   & Chin   \\ \hline
Deepfakes      & 0.0217 & 0.0218 & 0.0479 & 0.0325 \\
Face2Face      & 0.0218 & 0.0206 & 0.0450 & 0.0371 \\
FaceSwap       & 0.0206 & 0.0239 & 0.0470 & 0.0313 \\
NeuralTextures & 0.0196 & 0.0239 & 0.0469 & 0.0410
\end{tabular}
\caption{Average normalized absolute difference between real and manipulated images, broken down by region of the face and manipulation algorithm used. We see that most algorithms have somewhat similar distributions, with the largest changes occurring in the eyes and chin.}
\label{tbl:diff}
\end{table}

On the other hand, where distributions are very different, we see more discrepancy in performance. For instance, between Face2Face and Deepfakes we see very different distributions around the mouth regions, where Face2Face has a higher peak and Deepfakes is flatter. This may reflect the poor transfer performance of the mouth-based detector when trained on Face2Face and evaluated on Deepfakes (.53 vs a .7 baseline for truncated Xception). 

Finally, when we look at Table \ref{tbl:diff} and observe the average absolute differences within regions, we see other distinct patterns. The most obvious pattern is that different parts have similar amounts of changes even between splits - there is far more variation between parts than between algorithms, with eyes changing the most while the nose and mouth change the least. 

Beyond that, though, we observe other interesting patterns with respect to the performance of parts detectors. For instance, we see that  in the NeuralTextures split the chin region changes far more than in the other splits, and in fact in Table \ref{tbl:parts} we see that the chin-based detector trained on NeuralTextures considerably out-performs the baseline in the Deepfake and Face2Face dataset splits (.84 and .68 vs .69 and .55, respectively). The one split where the chin-based detector consistently under-performs the baseline is FaceSwap (.38 vs .42), the split with the smallest changes - and therefore presumably the fewest artifacts - in the chin region.

\subsubsection{Learned Masks}

\begin{figure*}[]
\begin{center}
   \includegraphics[width=0.95\linewidth]{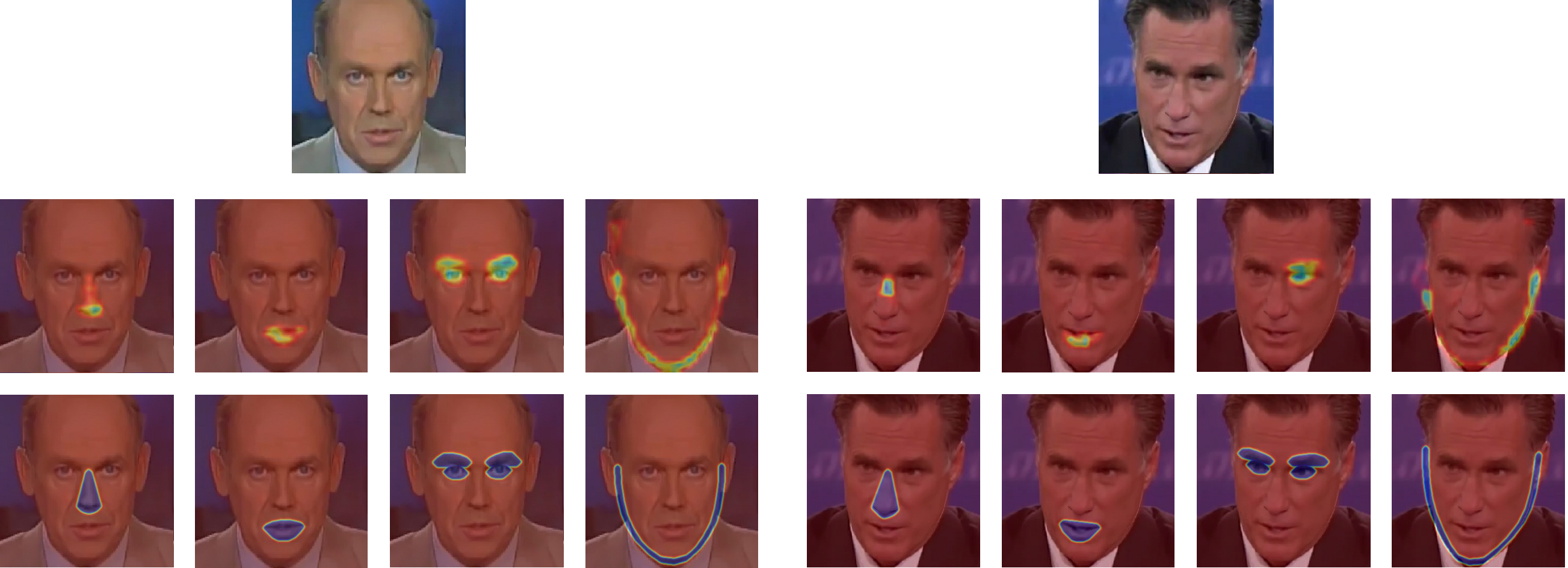}
\end{center}
   \caption{Masks learned by individual part detectors. The top row shows the masks learned by the various parts detectors, while the bottom row shows the ground truth masks constructed as described in Section \ref{sec:mask_gen} Left: a manipulated image from the Deepfakes FaceForensics++ split, evaluated by parts-based models trained on the Deepfakes split. Right: A manipulated image from the Face2Face split of FaceForensics++, evaluated on models trained on the Deepfakes split. }
\label{fig:masks}
\end{figure*}

One advantage to using parts-based classifiers is their explainability. When a parts-based system identifies an image or video as real or fake, one need only look at the predicted masks (described in Section \ref{sec:arch_single}) for hints at what the system was used to make its prediction. In Figure \ref{fig:masks} we provide two examples of these masks for each of the four facial regions. On the left we have the relatively easy scenario of identifying a fake from the same distribution as the network was trained on, in this case the Deepfakes split of FaceForensics++. We see that the system easily identifies all of the relevant regions as fakes.

On the right side of Figure \ref{fig:masks}, we have a sample of masks generated by parts-based detectors trained on the Deepfakes split and evaluated on Face2Face. Even transferring between splits, we see that the network is still able to find enough artifacts of the forgery process to be effective. For instance, the detector is still able to successfully pick up the small discoloration on the individual's nose, even if the masks are less clean and more likely to miss certain regions, such as the left eye.

\subsubsection{Architecture Design}

\begin{table}[]
\centering
\begin{tabular}{l|llll}
Model                & DF    & F2F   & FS    & NT    \\ \hline
Mean                 & 0.601 & 0.486 & 0.643 & 0.597 \\
Max                  & 0.645 & 0.529 & 0.58  & 0.571 \\
FC                   & 0.573 & 0.519 & 0.583 & 0.555 \\
Ensemble   & 0.587 & 0.462 & 0.576 & 0.61 
\end{tabular}
\caption{Different aggregation methods for the composite parts-based model. ``Mean'' is the method used in the rest of the paper which performs average pooling, ``Max'' performs max pooling, ``FC'' adds a fully connected layer, and ``Ensemble'' runs a separate network for each part, averaging the final logits.}
\label{tbl:agg}
\end{table}

\textbf{Choice of Aggregation Function} In all of the experiments we explored above, we chose to use an average pooling layer for aggregation in order to aggregate the results of our multi-headed parts detector. In Table \ref{tbl:agg}, we explore the choice of other aggregation methods, including a max pooling layer and a trained fully connected layer. These layers are all applied after the individual parts-maps are aggregated separately in the spatial domain using an average pooling layer. We also include one additional method of aggregation  labeled ``Ensemble'', which is attained by training each part detector with a completely different network, and then averaging the logits in an ensemble. Though this would of course be less practical since it involves training and running multiple networks, and the performance gains do not make it worthwhile, we do note that the final ensemble has only 1.6 million parameters and uses 14.6 billion FLOPS, which is still less than a full Xception architecture containing over 20 million parameters and using over 15 billion FLOPS.

In order to compute the values in Table \ref{tbl:agg}, we train and evaluate each of our architectures on the same sixteen combinations of splits from the FaceForensics++ dataset that we used in Table \ref{tbl:parts}. We then average all the results for a given split, ignoring results trained and evaluated on the same split. For example, for the Deepfakes split, we average the AUC results on the Deepfakes split for the architecture trained on Face2Face, FaceSwap, and NeuralTextures. This provides us with a good summary statistic for comparing the performance of all three architectural options. The full values obtained by all of these runs are included in the supplementary material.

We find that average and max pooling perform similarly, each out-performing the other in two out of the four categories. The fully-connected layer is generally an inferior form of aggregation, perhaps because it can encode biases for one part or another that do not transfer well between dataset splits.

\begin{table}[]
\centering
\begin{tabular}{l|llll}
Model    & DF    & F2F   & FS    & NT    \\ \hline
0 Blocks & 0.528 & 0.446 & 0.523 & 0.56  \\
1 Block  & 0.601 & 0.486 & 0.643 & 0.597 \\
2 Blocks & 0.575 & 0.497 & 0.582 & 0.565
\end{tabular}
\caption{Performance of the aggregated parts-based model, trained with different numbers of Xception blocks included after truncation.}
\label{tbl:blocks}
\end{table}

\textbf{Number of Additional Blocks} Another axis of variation in our architecture is the choice of the number of additional Xception blocks used. Adding these extra blocks is an essential step, because they allow the different part detectors to operate separately. We explore this in Table \ref{tbl:blocks}, where we have trained the combined architecture using average-pooling aggregation over 0, 1, and 2 additional Xception blocks. Here, we aggregate values in the same manner as we did in Table \ref{tbl:agg}, and once again note that the full experiments can be found in the supplementary material.  From this analysis, we find that adding one additional Xception block after truncation is optimal in most cases, with only a small loss with respect to the Face2Face split (from .497 to .486), which is easily outweighed by the larger gains made in the other three splits. The poor performance of using zero blocks in particular confirms our hypothesis that it is necessary to have at least some degree of separate processing for the individual parts-based models, as trying to detect all parts in a single branch will cause the detectors to interfere with one another and performance will degrade.

\subsubsection{Failure Case Analysis}

\begin{figure}[]
\begin{center}
   \includegraphics[width=0.65\linewidth]{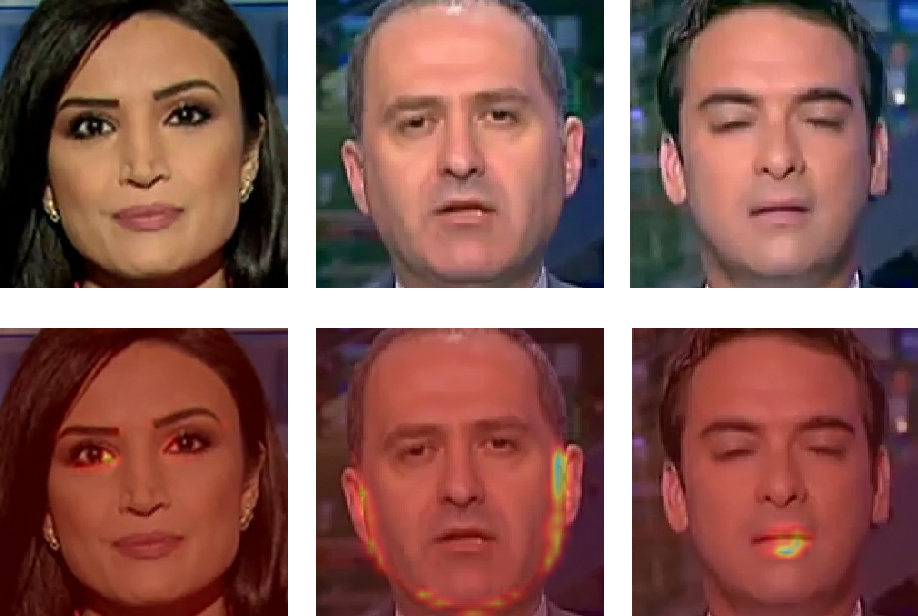}
\end{center}
   \caption{False positives for the eyes (left), chin (middle), and mouth (right) detectors. All three are incorrectly firing on regions within real images.}
\label{fig:fail}
\end{figure}

We can also find new insights by looking at the failure cases for a given parts-detector. Figure \ref{fig:fail} shows some examples of false positives detected by our system for different part-based detectors. More such false positives can also be found in the supplementary material. While we find that these failures look generally how we would expect them to, showing activations in areas where they should not exist, for combined parts detectors the ability to observe these maps when the algorithm fails would help an observer understand which regions of the image caused the network to trigger incorrectly.

\subsection{Cross-Dataset Generalization}

\begin{table}[]
\centering
\begin{tabular}{l|lll}
Model            & FF++                         & Celeb-DF       & DFDC           \\ \hline
Xception         & {\color[HTML]{9B9B9B} 0.965} & 0.629          & \textbf{0.673} \\
Xception Block 2 & {\color[HTML]{9B9B9B} 0.754} & 0.622          & 0.593          \\ \hdashline
Nose             & {\color[HTML]{9B9B9B} 0.909} & \textbf{0.667} & 0.611          \\
Mouth            & {\color[HTML]{9B9B9B} 0.914} & 0.658          & 0.617          \\
Eyes             & {\color[HTML]{9B9B9B} 0.847} & 0.63           & 0.586          \\
Chin             & {\color[HTML]{9B9B9B} 0.92}  & 0.644          & 0.618          \\
Average          & {\color[HTML]{9B9B9B} 0.931} & 0.633          & 0.627         
\end{tabular}
\caption{Performance of various systems when trained on the entire FaceForensics++ dataset and evaluated on the Celeb-DF and Facebook DFDC datasets. Our parts-based detectors are below the dotted line. Best results are in bold.}
\label{tbl:transfer}
\end{table}

We have shown that parts-based detectors generalize well between splits of the FaceForensics++ datasets. In this section, we evaluate the generalization performance of parts-based detectors trained on the entirety of the FaceFroensics dataset and evaluated on two other datasets, Celeb-DF and Facebook DFDC. For both of these datasets, our evaluation is only on the publicly-available test splits.

The results of these experiments are shown in Table \ref{tbl:transfer}, using the same AUC metric used above. We observe that, with respect to the Celeb-DF dataset, parts-based detectors are generally superior to the Xception baselines, both truncated and otherwise. This shows that our method remains effective relative to other techniques even as the difference between training and evaluation grows.

However, for the DFDC dataset, we find that parts detectors are not sufficient to achieve state-of-the-art performance over the Xception baseline. However, we still observe that all parts-based methods out-perform the truncated Xception baseline, which itself outperformed the Xception baseline in the vast majority of other transfer tasks. Overall, this indicates that some of the data in the DFDC dataset might require a more global approach in order to perform proper detection, whereas more local approaches with smaller receptive fields, such as truncated Xception and parts-based classifiers, simply do not have sufficient receptive fields. This also opens up the possibility of future work into parts-based detectors that use much larger receptive fields, perhaps with variants of U-Net \cite{unet} or similar architectures.

\section{Conclusion}

In this work, we have shown that it is possible to use neural networks trained only to look in specific regions of the face to improve generalization performance between video manipulation algorithms. This suggests that these individual parts of the face may in some cases actually be more representative than the rest of the image as a whole, since restricting the classifier's attention to only these parts improves generalization. Having observed this, we used these parts-based detectors to perform extensive empirical analysis, analyzing which parts are most discriminative between datasets and examining the underlying distributions of our data. 

\section{Acknowledgements}

This paper is based upon work supported, in whole or in part, by the Office of the Under Secretary of Defense for Intelligence and Security. Any opinions, findings and conclusions or recommendations expressed in this material are those of the author(s) and do not necessarily reflect the views of the University of Maryland, the UMD Applied Research Laboratory for Intelligence and Security (ARLIS), Office of the Under Secretary of Defense for Intelligence and Security, or the U.S. Government.

{\small
\bibliographystyle{ieee_fullname}
\bibliography{egbib}
}

\section{Supplementary Material}

\subsection{Full Data}

Here we include more complete data from our experiments. Table \ref{tbl:full} shows complete data for Table 1 in the full paper, Table \ref{tbl:methods} shows complete data for Table 3 in the full paper, and Table \ref{tbl:blocks2} shows complete data for Table 4 in the original paper.

For Table \ref{tbl:full}, the key difference is in the addition of the accuracy metric, computed as the average accuracy of real and fake computation. Since for each split there are the same number of real and fake images, the metric does not need to be re-balanced. Tables \ref{tbl:methods} and \ref{tbl:blocks2} contain the accuracy metric as well, but also include detailed AUC numbers per split which were not included in the full paper.

\begin{table*}[]
\centering
\begin{tabular}{ll|llllllll}
Model            &     & DF Acc                      & F2F Acc                     & FS Acc                      & NT Acc                      & DF AUC                      & F2F AUC                     & FS AUC                      & NT AUC                      \\ \hline
ResNet50         & DF  & {\color[HTML]{9B9B9B} 0.96} & 0.5                         & 0.48                        & 0.55                        & {\color[HTML]{9B9B9B} 0.98} & 0.51                        & 0.48                        & 0.57                        \\
ResNet50 Block 2 & DF  & {\color[HTML]{9B9B9B} 0.95} & 0.5                         & 0.5                         & 0.54                        & {\color[HTML]{9B9B9B} 0.99} & 0.55                        & 0.47                        & 0.68                        \\
Xception         & DF  & {\color[HTML]{9B9B9B} 0.97} & 0.51                        & 0.5                         & 0.58                        & {\color[HTML]{9B9B9B} 0.98} & 0.52                        & 0.49                        & 0.61                        \\
Xception Block 2 & DF  & {\color[HTML]{9B9B9B} 0.87} & 0.54                        & 0.45                        & 0.72                        & {\color[HTML]{9B9B9B} 0.91} & 0.6                         & 0.43                        & 0.79                        \\ \hdashline
Nose             & DF  & {\color[HTML]{9B9B9B} 0.87} & 0.57                        & 0.41                        & 0.74                        & {\color[HTML]{9B9B9B} 0.97} & 0.63                        & 0.33                        & 0.81                        \\
Mouth            & DF  & {\color[HTML]{9B9B9B} 0.82} & 0.55                        & 0.44                        & 0.7                         & {\color[HTML]{9B9B9B} 0.94} & 0.56                        & 0.48                        & 0.76                        \\
Eyes             & DF  & {\color[HTML]{9B9B9B} 0.88} & 0.54                        & 0.49                        & 0.74                        & {\color[HTML]{9B9B9B} 0.96} & 0.6                         & 0.4                         & 0.84                        \\
Chin             & DF  & {\color[HTML]{9B9B9B} 0.89} & 0.52                        & 0.48                        & 0.75                        & {\color[HTML]{9B9B9B} 0.96} & 0.59                        & 0.33                        & 0.85                        \\
Eyes+Chin        & DF  & {\color[HTML]{9B9B9B} 0.92} & 0.52                        & 0.49                        & 0.67                        & {\color[HTML]{9B9B9B} 0.97} & 0.58                        & 0.42                        & 0.76                        \\
Combined         & DF  & {\color[HTML]{9B9B9B} 0.9}  & 0.51                        & 0.5                         & 0.67                        & {\color[HTML]{9B9B9B} 0.97} & 0.62                        & 0.39                        & 0.83                        \\ \hline
ResNet50         & F2F & 0.56                        & {\color[HTML]{9B9B9B} 0.97} & 0.51                        & 0.53                        & 0.57                        & {\color[HTML]{9B9B9B} 0.98} & 0.5                         & 0.56                        \\
ResNet50 Block 2 & F2F & 0.53                        & {\color[HTML]{9B9B9B} 0.98} & 0.51                        & 0.52                        & 0.66                        & {\color[HTML]{9B9B9B} 0.99} & 0.54                        & 0.65                        \\
Xception         & F2F & 0.53                        & {\color[HTML]{9B9B9B} 0.98} & 0.5                         & 0.51                        & 0.58                        & {\color[HTML]{9B9B9B} 0.98} & 0.52                        & 0.54                        \\
Xception Block 2 & F2F & 0.66                        & {\color[HTML]{9B9B9B} 0.71} & 0.63                        & 0.68                        & 0.7                         & {\color[HTML]{9B9B9B} 0.94} & 0.64                        & 0.74                        \\ \hdashline
Nose             & F2F & 0.54                        & {\color[HTML]{9B9B9B} 0.96} & 0.53                        & 0.53                        & 0.62                        & {\color[HTML]{9B9B9B} 0.98} & 0.58                        & 0.57                        \\
Mouth            & F2F & 0.51                        & {\color[HTML]{9B9B9B} 0.94} & 0.55                        & 0.49                        & 0.53                        & {\color[HTML]{9B9B9B} 0.98} & 0.65                        & 0.52                        \\
Eyes             & F2F & 0.6                         & {\color[HTML]{9B9B9B} 0.93} & 0.58                        & 0.58                        & 0.73                        & {\color[HTML]{9B9B9B} 0.97} & 0.65                        & 0.72                        \\
Chin             & F2F & 0.59                        & {\color[HTML]{9B9B9B} 0.89} & 0.56                        & 0.58                        & 0.75                        & {\color[HTML]{9B9B9B} 0.95} & 0.65                        & 0.74                        \\
Eyes+Chin        & F2F & 0.5                         & {\color[HTML]{9B9B9B} 0.94} & 0.53                        & 0.48                        & 0.56                        & {\color[HTML]{9B9B9B} 0.98} & 0.62                        & 0.48                        \\
Combined         & F2F & 0.68                        & {\color[HTML]{9B9B9B} 0.87} & 0.52                        & 0.71                        & 0.76                        & {\color[HTML]{9B9B9B} 0.95} & 0.53                        & 0.77                        \\ \hline
ResNet50         & FS  & 0.5                         & 0.52                        & {\color[HTML]{9B9B9B} 0.97} & 0.49                        & 0.5                         & 0.54                        & {\color[HTML]{9B9B9B} 0.99} & 0.49                        \\
ResNet50 Block 2 & FS  & 0.51                        & 0.52                        & {\color[HTML]{9B9B9B} 0.97} & 0.49                        & 0.54                        & 0.65                        & {\color[HTML]{9B9B9B} 1}    & 0.42                        \\
Xception         & FS  & 0.5                         & 0.54                        & {\color[HTML]{9B9B9B} 0.98} & 0.5                         & 0.51                        & 0.58                        & {\color[HTML]{9B9B9B} 0.99} & 0.5                         \\
Xception Block 2 & FS  & 0.48                        & 0.58                        & {\color[HTML]{9B9B9B} 0.88} & 0.45                        & 0.48                        & 0.62                        & {\color[HTML]{9B9B9B} 0.92} & 0.29                        \\ \hdashline
Nose             & FS  & 0.51                        & 0.53                        & {\color[HTML]{9B9B9B} 0.96} & 0.49                        & 0.53                        & 0.56                        & {\color[HTML]{9B9B9B} 0.98} & 0.47                        \\
Mouth            & FS  & 0.48                        & 0.59                        & {\color[HTML]{9B9B9B} 0.8}  & 0.34                        & 0.45                        & 0.63                        & {\color[HTML]{9B9B9B} 0.96} & 0.22                        \\
Eyes             & FS  & 0.51                        & 0.52                        & {\color[HTML]{9B9B9B} 0.96} & 0.49                        & 0.53                        & 0.64                        & {\color[HTML]{9B9B9B} 0.98} & 0.45                        \\
Chin             & FS  & 0.49                        & 0.52                        & {\color[HTML]{9B9B9B} 0.95} & 0.49                        & 0.44                        & 0.73                        & {\color[HTML]{9B9B9B} 0.99} & 0.33                        \\
Eyes+Chin        & FS  & 0.5                         & 0.54                        & {\color[HTML]{9B9B9B} 0.92} & 0.49                        & 0.48                        & 0.71                        & {\color[HTML]{9B9B9B} 0.99} & 0.41                        \\
Combined         & FS  & 0.54                        & 0.55                        & {\color[HTML]{9B9B9B} 0.93} & 0.44                        & 0.52                        & 0.56                        & {\color[HTML]{9B9B9B} 0.98} & 0.32                        \\ \hline
ResNet50         & NT  & 0.55                        & 0.5                         & 0.49                        & {\color[HTML]{9B9B9B} 0.93} & 0.56                        & 0.51                        & 0.48                        & {\color[HTML]{9B9B9B} 0.94} \\
ResNet50 Block 2 & NT  & 0.55                        & 0.5                         & 0.5                         & {\color[HTML]{9B9B9B} 0.96} & 0.67                        & 0.52                        & 0.43                        & {\color[HTML]{9B9B9B} 0.98} \\
Xception         & NT  & 0.55                        & 0.57                        & 0.5                         & {\color[HTML]{9B9B9B} 0.98} & 0.59                        & 0.6                         & 0.5                         & {\color[HTML]{9B9B9B} 1}    \\
Xception Block 2 & NT  & 0.53                        & 0.5                         & 0.5                         & {\color[HTML]{9B9B9B} 0.93} & 0.69                        & 0.55                        & 0.42                        & {\color[HTML]{9B9B9B} 0.98} \\ \hdashline
Nose             & NT  & 0.52                        & 0.51                        & 0.5                         & {\color[HTML]{9B9B9B} 0.98} & 0.54                        & 0.51                        & 0.5                         & {\color[HTML]{9B9B9B} 0.98} \\
Mouth            & NT  & 0.52                        & 0.51                        & 0.5                         & {\color[HTML]{9B9B9B} 0.95} & 0.64                        & 0.63                        & 0.54                        & {\color[HTML]{9B9B9B} 0.99} \\
Eyes             & NT  & 0.54                        & 0.5                         & 0.5                         & {\color[HTML]{9B9B9B} 0.97} & 0.64                        & 0.53                        & 0.47                        & {\color[HTML]{9B9B9B} 0.99} \\
Chin             & NT  & 0.67                        & 0.54                        & 0.49                        & {\color[HTML]{9B9B9B} 0.97} & 0.84                        & 0.68                        & 0.38                        & {\color[HTML]{9B9B9B} 0.99} \\
Eyes+Chin        & NT  & 0.57                        & 0.52                        & 0.5                         & {\color[HTML]{9B9B9B} 0.97} & 0.77                        & 0.61                        & 0.47                        & {\color[HTML]{9B9B9B} 0.98} \\
Combined         & NT  & 0.52                        & 0.5                         & 0.5                         & {\color[HTML]{9B9B9B} 0.96} & 0.66                        & 0.62                        & 0.53                        & {\color[HTML]{9B9B9B} 0.98}
\end{tabular}
\caption{AUC for the ROC curves of the parts-based detectors for each of the four parts of the face, as well as for the combined detector. The second column indicates which split of FaceForensics++ was used to train the model, while the other columns show the performance on each of those splits. This is an extension of Table 1 in the full paper. Here we include both standard accuracy (percentage of correct detections in each class) and AUC of the ROC curve.}
\label{tbl:full}
\end{table*}

\begin{table*}[]
\centering
\begin{tabular}{ll|llllllll}
Model    &     & DF Acc                      & F2F Acc                     & FS Acc                      & NT Acc                      & DF AUC                      & F2F AUC                     & FS AUC                      & NT AUC                      \\ \hline
FC       & DF  & {\color[HTML]{9B9B9B} 0.92} & 0.51                        & 0.49                        & 0.56                        & {\color[HTML]{9B9B9B} 0.98} & 0.57                        & 0.49                        & 0.63                        \\
Mean     & DF  & {\color[HTML]{9B9B9B} 0.9}  & 0.51                        & 0.5                         & 0.67                        & {\color[HTML]{9B9B9B} 0.97} & 0.62                        & 0.39                        & 0.83                        \\
Max      & DF  & {\color[HTML]{9B9B9B} 0.89} & 0.54                        & 0.47                        & 0.67                        & {\color[HTML]{9B9B9B} 0.96} & 0.58                        & 0.45                        & 0.73                        \\
Ensemble & DF  & {\color[HTML]{9B9B9B} 0.93} & 0.53                        & 0.46                        & 0.72                        & {\color[HTML]{9B9B9B} 0.98} & 0.61                        & 0.41                        & 0.83                        \\ \hline
FC       & F2F & 0.63                        & {\color[HTML]{9B9B9B} 0.83} & 0.58                        & 0.67                        & 0.66                        & {\color[HTML]{9B9B9B} 0.97} & 0.59                        & 0.69                        \\
Mean     & F2F & 0.68                        & {\color[HTML]{9B9B9B} 0.87} & 0.52                        & 0.71                        & 0.76                        & {\color[HTML]{9B9B9B} 0.95} & 0.53                        & 0.77                        \\
Max      & F2F & 0.63                        & {\color[HTML]{9B9B9B} 0.87} & 0.62                        & 0.56                        & 0.72                        & {\color[HTML]{9B9B9B} 0.96} & 0.69                        & 0.65                        \\
Ensemble & F2F & 0.54                        & {\color[HTML]{9B9B9B} 0.96} & 0.53                        & 0.53                        & 0.77                        & {\color[HTML]{9B9B9B} 0.99} & 0.68                        & 0.74                        \\ \hline
FC       & FS  & 0.51                        & 0.56                        & {\color[HTML]{9B9B9B} 0.95} & 0.48                        & 0.49                        & 0.59                        & {\color[HTML]{9B9B9B} 0.98} & 0.43                        \\
Mean     & FS  & 0.54                        & 0.55                        & {\color[HTML]{9B9B9B} 0.93} & 0.44                        & 0.52                        & 0.56                        & {\color[HTML]{9B9B9B} 0.98} & 0.32                        \\
Max      & FS  & 0.5                         & 0.55                        & {\color[HTML]{9B9B9B} 0.96} & 0.49                        & 0.5                         & 0.73                        & {\color[HTML]{9B9B9B} 0.99} & 0.36                        \\
Ensemble & FS  & 0.51                        & 0.53                        & {\color[HTML]{9B9B9B} 0.96} & 0.49                        & 0.5                         & 0.62                        & {\color[HTML]{9B9B9B} 0.99} & 0.33                        \\ \hline
FC       & NT  & 0.55                        & 0.51                        & 0.49                        & {\color[HTML]{9B9B9B} 0.95} & 0.59                        & 0.55                        & 0.48                        & {\color[HTML]{9B9B9B} 0.98} \\
Mean     & NT  & 0.52                        & 0.5                         & 0.5                         & {\color[HTML]{9B9B9B} 0.96} & 0.66                        & 0.62                        & 0.53                        & {\color[HTML]{9B9B9B} 0.98} \\
Max      & NT  & 0.64                        & 0.53                        & 0.5                         & {\color[HTML]{9B9B9B} 0.97} & 0.77                        & 0.63                        & 0.45                        & {\color[HTML]{9B9B9B} 0.98} \\
Ensemble & NT  & 0.52                        & 0.51                        & 0.5                         & {\color[HTML]{9B9B9B} 0.98} & 0.76                        & 0.6                         & 0.42                        & {\color[HTML]{9B9B9B} 0.99}
\end{tabular}
\caption{Different aggregation methods for the composite parts-based model. ``Mean'' is the method used in the rest of the paper which performs average pooling, ``Max'' performs max pooling, ``FC'' adds a fully connected layer, and ``Ensemble'' runs a separate network for each part, averaging the final logits. This is an extension of Table 3 in the full paper. Here we include both standard accuracy (percentage of correct detections in each class) and AUC of the ROC curve.}
\label{tbl:methods}
\end{table*}

\begin{table*}[]
\centering
\begin{tabular}{ll|llllllll}
Model    &     & DF Acc                      & F2F Acc                     & FS Acc                      & NT Acc                      & DF AUC                      & F2F AUC                     & FS AUC                      & NT AUC                      \\ \hline
0 Blocks & DF  & {\color[HTML]{9B9B9B} 0.83} & 0.54                        & 0.42                        & 0.68                        & {\color[HTML]{9B9B9B} 0.93} & 0.55                        & 0.32                        & 0.75                        \\
1 Block  & DF  & {\color[HTML]{9B9B9B} 0.9}  & 0.51                        & 0.5                         & 0.67                        & {\color[HTML]{9B9B9B} 0.97} & 0.62                        & 0.39                        & 0.83                        \\
2 Blocks & DF  & {\color[HTML]{9B9B9B} 0.91} & 0.53                        & 0.46                        & 0.7                         & {\color[HTML]{9B9B9B} 0.96} & 0.6                         & 0.41                        & 0.79                        \\ \hline
0 Blocks & F2F & 0.52                        & {\color[HTML]{9B9B9B} 0.53} & 0.5                         & 0.55                        & 0.52                        & {\color[HTML]{9B9B9B} 0.53} & 0.5                         & 0.55                        \\
1 Block  & F2F & 0.68                        & {\color[HTML]{9B9B9B} 0.87} & 0.52                        & 0.71                        & 0.76                        & {\color[HTML]{9B9B9B} 0.95} & 0.53                        & 0.77                        \\
2 Blocks & F2F & 0.52                        & {\color[HTML]{9B9B9B} 0.94} & 0.52                        & 0.5                         & 0.6                         & {\color[HTML]{9B9B9B} 0.99} & 0.6                         & 0.63                        \\ \hline
0 Blocks & FS  & 0.52                        & 0.5                         & {\color[HTML]{9B9B9B} 0.87} & 0.44                        & 0.49                        & 0.5                         & {\color[HTML]{9B9B9B} 0.93} & 0.27                        \\
1 Block  & FS  & 0.54                        & 0.55                        & {\color[HTML]{9B9B9B} 0.93} & 0.44                        & 0.52                        & 0.56                        & {\color[HTML]{9B9B9B} 0.98} & 0.32                        \\
2 Blocks & FS  & 0.52                        & 0.55                        & {\color[HTML]{9B9B9B} 0.91} & 0.43                        & 0.49                        & 0.58                        & {\color[HTML]{9B9B9B} 0.99} & 0.32                        \\ \hline
0 Blocks & NT  & 0.52                        & 0.51                        & 0.52                        & {\color[HTML]{9B9B9B} 0.84} & 0.55                        & 0.53                        & 0.52                        & {\color[HTML]{9B9B9B} 0.92} \\
1 Block  & NT  & 0.52                        & 0.5                         & 0.5                         & {\color[HTML]{9B9B9B} 0.96} & 0.66                        & 0.62                        & 0.53                        & {\color[HTML]{9B9B9B} 0.98} \\
2 Blocks & NT  & 0.51                        & 0.5                         & 0.5                         & {\color[HTML]{9B9B9B} 0.95} & 0.58                        & 0.54                        & 0.48                        & {\color[HTML]{9B9B9B} 0.98}
\end{tabular}
\caption{Performance of the aggregated parts-based model, trained with different numbers of Xception blocks included after truncation. This is an extension of Table 4 in the full paper. Here we include both standard accuracy (percentage of correct detections in each class) and AUC of the ROC curve.}
\label{tbl:blocks2}
\end{table*}



\end{document}